\documentclass[10pt,twocolumn]{article}

\usepackage{iccv}
\usepackage{times}
\usepackage{epsfig}
\usepackage{graphicx}
\usepackage{amsmath}
\usepackage{amssymb}

\usepackage{latexsym,color,graphicx,amsmath,amssymb,subcaption}
\usepackage{url}

\usepackage{latexsym,color,graphicx,amsmath,amssymb,subcaption}
\usepackage{float}
\usepackage{fancyhdr}
\usepackage[ruled,vlined]{algorithm2e}
\usepackage{complexity}
\usepackage{graphicx}
\graphicspath{{figures/}}
\usepackage{comment}
\usepackage{xfrac}   
\usepackage{booktabs}
\usepackage{multirow}

\usepackage{enumitem}
\setitemize{noitemsep,topsep=0pt,parsep=0pt,partopsep=0pt}

\DeclareMathOperator*{\argmax}{argmax}
\def\reals{{\mathbb R}}
\def\eps{{\varepsilon}}

\def\ww{{\bf w}}
\def\xx{{\bf x}}

\def\cc{{\bf c}}

\def\ff{{\bf f}}
\def\g{{\bf g}}

\def\polylog{\operatorname{polylog}}

\usepackage[pagebackref=true,breaklinks=true,letterpaper=true,colorlinks,bookmarks=false]{hyperref}

\iccvfinalcopy 

\def\iccvPaperID{10801} 

\ificcvfinal\fi

\begin{document}

\title{Efficient Large Scale Inlier Voting for Geometric Vision Problems}

\author{Dror Aiger\\
Google\\
{\tt\small aigerd@google.com}
\and
Simon Lynen\\
Google\\
{\tt\small slynen@google.com}
\and
Jan Hosang\\
Google\\
{\tt\small hosang@google.com}
\and
Bernhard Zeisl\\
Google\\
{\tt\small bzeisl@google.com}
}

\maketitle

\begin{abstract}
Outlier rejection and equivalently inlier set optimization is a key ingredient in numerous applications in computer vision such as filtering point-matches in camera pose estimation or plane and normal estimation in point clouds.
Several approaches exist, yet at large scale we face a combinatorial explosion of possible solutions and state-of-the-art methods like RANSAC, Hough transform or Branch\&Bound require a minimum inlier ratio or prior knowledge to remain practical.
In fact, for problems such as camera posing in very large scenes these approaches become useless as they have exponential runtime growth if these conditions aren't met.
%
%

To approach the problem we present a efficient and general algorithm for outlier rejection based on ``intersecting" $k$-dimensional surfaces in $R^d$.
We provide a recipe for casting a variety of geometric problems as finding a point in $R^d$ which maximizes the number of nearby surfaces (and thus inliers).
%
%
The resulting algorithm has linear worst-case complexity with a better runtime dependency in the approximation factor than competing algorithms while not requiring domain specific bounds.
This is achieved by introducing a space decomposition scheme that bounds the number of computations by successively rounding and grouping samples.
Our recipe (and open-source code) enables anybody to derive such fast approaches to new problems across a wide range of domains.
We demonstrate the versatility of the approach on several camera posing problems with a high number of matches at low inlier ratio achieving state-of-the-art results at significantly lower processing times.

\end{abstract}

\section{Introduction}
%
%
Whether geometric verification of point matches for absolute pose, homography estimation or normal estimation in point clouds, outlier rejection is a key ingredient in numerous applications in computer vision.
%
%
RANSAC~\cite{FB81} is a common choice for the task as it has good quality/runtime tradeoffs for many practical problems.
RANSAC however doesn't provide an optimality guarantee, requires a lower bound on the inlier ratio and its runtime increases exponentially with the outlier ratio making it unusable for problems with few inliers.
In contrast Branch\&Bound methods~\cite{B03,CPKH20} are guaranteed to find the optimal solution, yet their worst-case runtime equals exhaustive search in parameter space. Their practical use depends on the quality of the bounds which are problem specific and notoriously hard to find.

%
We propose an alternative approach with linear runtime independent of the inlier ratio and that doesn't require a lower bound on the fraction of inliers. 
%
We base our work on the efficient general algorithm for solving geometric incidence problems proposed in \cite{AKKSZ19}.
In a nutshell the authors show that voting \cite{ZSP15} is equivalent to finding a point of maximum depth, i.e. to find a point in $\reals^d$ which is close to as many surfaces (and thus inliers) as possible among a given set of $k$-surfaces in $\reals^d$.
We demonstrate an effective solution to outlier removal problems by transforming them to the surface incidence framework of~\cite{AKKSZ19}.
Our contributions:


\begin{itemize}
    \item[--] We introduce the concept of \textit{general voting} and its relation to the \textit{approximate incidences} of~\cite{AKKSZ19} for outlier removal to a wider computer vision audience and demonstrate its use for camera posing, ray intersection and geometric model fitting.
    \item[--] All approaches have \textit{linear} complexity and are therefore applicable to very large problems, which are infeasible to solve with RANSAC~\cite{FB81}. The worst case complexity (and performance in practice) is always better than other methods (voting~\cite{HOU62} and B\&B~\cite{HD60}) for sufficiently large inputs.
    \item[--] We demonstrate the scalability and versatility using a generalization of \cite{ZSP15,SEKO16} where we remove the requirement on a calibrated camera and known gravity direction, but instead solve for these unknowns.
    \item[--] We compare to the state-of-the-art~\cite{CPKH20} for 6 degree-of-freedom (6DoF) camera pose estimation and show that our approach is optimal and faster in practice. In comparison to~\cite{CPKH20} our surface intersection algorithm provides a tight upper bound on the score without problem specific knowledge.
    \item[--] We provide example-derivations and open-source code which serve as a tutorial for applying~\cite{AKKSZ19} to a range of outlier removal problems in computer vision.
\end{itemize}

\section{The Family of General Alignment Problems}

A large number of geometric vision problems can be viewed as general alignment problems where we aim to bring items in set a $A$ "close" to items in other set $B$ by applying a transformation from a group of allowed transformations.
This closeness can be defined by some given parameter, $\eps>0$ and it is common to use the Hausdorff~\cite{HK90} metric, the distance between items in $A$ and $B$ w.r.t. $\eps$.

This alignment problem is at the core of many outlier rejection problems (e.g. the ones discussed initially).
There we aim to bring a maximum number of items in $A$ to be at Hausdorff distance at most $\eps$ from some item in $B$:
For example in structure-based localization~\cite{LSHF12,SMTTHSSOPS18} $A$ is a set of rays in $3$-space in the camera frame that we want to align with a set of $3$d world points $B$.
In relative camera posing~\cite{FLOEK16,FL16} we aim to align two sets of $3$-rays such that the maximum number of pairs approximately intersect $\eps$-close.
%
%
In all of these problems, one defines the objects and the allowed transformation group and then solves an approximate geometric incidence problem.

%
%

Any hypothetical match --- a correspondence between an object in $A$ and an object in $B$ --- defines a \textit{general surface} of some dimension which is embedded in the ambient $d$-space of transformations.
Therefore, we can solve the inlier set maximization in linear time in the number of hypothetical matches by some kind of voting scheme: surfaces are intersected in the $d$-space to locate the point closest to as many surfaces as possible.
This point of maximum incidences represents the solution of the alignment problem.
%

\section{Proximity and Incidences using Surfaces}
\label{inc_to_surf}
Our work is based on the findings of~\cite{AKKSZ19} and we introduce their approach and notation using 2D line fitting as a toy example.
The authors formulate the maximum incidence problem as one of reporting points in $\reals^d$ that are close to the majority of $k$-dimensional surfaces.
Each surface $\sigma \in S$ is given in parametric form where the first $k$ coordinates $\xx = (x_1,\ldots,x_k)$ are the surface parameters.
Specifically, each $\sigma$ is defined in terms of \emph{$\ell$ essential parameters} ${\bf t}=(t_1,\ldots,t_\ell)$, and $d-k$ additional \emph{free additive parameters} $\ff=(f_{k+1},\ldots,f_d)$, one free parameter for each dependent coordinate.
The surface $\sigma$ is parameterized by $\bf t$ and $\ff$ (we then denote $\sigma$ as $\sigma_{\bf t,\ff}$) and defined by
\begin{equation}
\begin{split}
& F^{(\sigma)}: \reals^k \times \reals^\ell \rightarrow \reals^{(d-k)}
\\
& x_j = F_j^{(\sigma)}(\xx; {\bf t}) + f_j, \qquad \text{for $j=k+1,\ldots,d$} .
\end{split}
\label{eq:surface}
\end{equation}

W.l.o.g.\ the voting space is scaled to be the unit cube $[0,1]^d$ to simplify runtime complexity expressions.

For our toy example, we are given a set of points in 2D to which we want to fit a 2D line.
Each  point is transformed, by a standard point-line duality~\cite{EA57} to a line, lines to points and for each point-line pair, the vertical distance between them is preserved. The $k=1$ dimensional surface embedded in the $\reals^{d=2}$ line parameter space, describing all lines that intersect the point. 
The (single) surface equation in this case is
\begin{equation}
x_2 =  a \cdot x_1 + b =  F_2^{(\sigma)}(x_1; a) + b,
\label{eq:line-fitting}
\end{equation}
where the slope $a$ of a line is the essential parameter $\bf t$ and the offset $b$ is the additive free parameter $\ff$.
Consequently, the point in $\reals^{d=2}$ with the largest number of close surfaces corresponds to the line parameters that fit the input points best (i.e.\ with maximal number of incidences or inliers).

\subsection{A Naive Voting Solution}
\label{sec:naive-voting}

Once we formulate a problem as a general surface consensus problem, we obtain a naive algorithm:
Iterate the first $k$ dimensions of the voting space on an $\eps$-grid and compute the dependent variables for each surface for an $3 \eps$ neighborhood box around each grid vertex.
This makes sure that for each vertex we collect all nearby surfaces at most $\eps$ apart from each other. Then it uses the dependent variable range to cast votes in the voting space (Algorithm~\ref{alg:naive-voting}).

For higher dimensional spaces this enumeration becomes costly, since the algorithm is independent of the actual structure of the surfaces and always takes $O(n / \eps^k)$ time for $n$ $k$-dimensional surfaces.


\begin{algorithm}
\SetAlgoLined
\DontPrintSemicolon
\KwData{$S$: surfaces, $B$: box in $[0,1]^d$, $\eps$: distance}
\KwResult{Point in $B$ (among all $\eps$-grid points) with the maximum of $\eps$-close surfaces.}
\For{all $s\in S$}{
 \For{all $k$-dimensional $\eps$-cells in $B$}{
    Compute the $d-k$ dependent variables using the surface equations.\;
    Tally a vote for the $d$-tuple and add $s$ into the set of inliers corresponding to it.\;
    /* The $d$-tuple is the concatenation of the $k$-tuple and the dependent $d-k$-tuple */
 }}
 \Return{The center of the cell with maximum votes and its corresponding set of intersecting surfaces.}\;
 \caption{Naive Voting} 
\label{alg:naive-voting}
\end{algorithm}

\subsection{Efficient Canonized Generalized Voting}
\label{sec:generalized-voting}
Algorithm~\ref{alg:generalized-voting} is our generalized voting procedure based on the work in~\cite{AKKSZ19}.
We propose a coarse-to-fine scheme that decomposes the search space to significantly improve the runtime complexity.
Our approach rounds surfaces so that we can group similar surfaces for joint processing, without affecting the outcome of the computation.
This canonization means that in every recursive step of the algorithm (similar to levels in an octree decomposition) there are approximately \emph{the same number of surfaces} to process.
The consequence is a worst case runtime complexity of $O(n+\sfrac{\polylog (1/\eps)}{\eps^{\ell+d-k}})$ (see Theorem~4.4 in~\cite{AKKSZ19}) or $O(n+\sfrac{1}{\eps^{\ell+d-k}})$ ignoring log factors.
For sufficiently large $n$ generalized voting is thus always asymptotically faster because of the multiplicative influence of $n$ on the approximation cost in naive voting.

For simplicity, we refer to a general $\eps$ parameter in the text.
For guaranteed approximation as in ~\cite{AKKSZ19} one needs to update $\eps$ slightly in order to compensate for accumulated error during the canonization and ensure that no close surface is missed.
In practice we use different $\eps$ for each coordinate and tune the required values experimentally.

\begin{algorithm}
\SetAlgoLined
\DontPrintSemicolon
\SetKwFunction{FnSurfaceConsensus}{SurfaceConsensus}
\SetKwProg{Fn}{Function}{:}{}
\KwData{$S$: surfaces, $B$: box in $[0,1]^d$, $\eps$: distance}
\KwResult{Point in $B$ (among all $\eps$-grid points) with the maximum of $\eps$-close surfaces.}
\Fn{\FnSurfaceConsensus{$S$, $B$, $\eps$}}{
\If{$Diam(B)\leq\eps$} {\Return{$(B_c,S)$} /* $B_c$ is the center of $B$ */}
 Canonize all surfaces $S$ (for $B$) to a new set $S_c$.\;
 Subdivide $B$ to $2^d$ sub-boxes, $B_i$.\;
 For each sub-box $B_i$, find a subset $S_{c_i} \subset S_c$ of surfaces that intersect it.\;
 \For{all $S_{c_i}$}{
    ($p_i,I_i$) = \FnSurfaceConsensus{$S_{c_i}$, $B_i$, $\eps$}\;
    /* $p_i$: a point in $B_i$, $I_i$: the set of inliers */
 }
 \Return{$(p_k=\argmax_{p_i} |I_i|, I_k)$}\;
 }
 \caption{Efficient Generalized Voting }
\label{alg:generalized-voting}
\end{algorithm}

\subsubsection{Algorithm and Implementation}
\label{algs}
Given a general outlier removal problem, the first step is to formulate the problem as a general surface with a parametric representation as in section~\ref{inc_to_surf}.

\vspace{\baselineskip}

\noindent\textbf{From constraints to general voting.}
We start with a set of constraints, each of which implicitly define a surface embedded in $R^d$ (by the points that satisfy the constraint). The surface may have only $k$ dimensions (where $k<=d$), meaning that w.l.o.g.\ given the other $(d-k)$ values, we can compute the point as a function of $(d-k)$ variables.
For 2D line fitting (Eq.~\eqref{eq:line-fitting}) every 2D point $(x_1,x_2)$ defines a line in voting space (with $k=1$, we have a $d-k=1$ dimensional surface embedded in the 2D ambient space).
To use our framework, we have to provide two functions: (1) A predicate that returns whether a given surface intersects a given box in $R^d$; (2) A function  $F^{(\sigma)}(\xx,{\bf t})+\ff$ that computes the $d-k$ dependent variables, given the other $k$ variables.
In our example of 2D line fitting (1) computes whether $ax_1+b$ intersects the given box $B$ and (2) is $x_2=ax_1+b$.

\vspace{\baselineskip}

\noindent\textbf{Canonization.} Most Branch\&Bound algorithms subdivide the parameter search, discarding branches early based on computed scores.
A common such subdivision is the octree.
A key ingredient that makes our generalized voting algorithm efficient is \emph{Canonization}; a surface rounding process we apply before recursing to the next level of the octree.
Surfaces which are close to each other in the current box are rounded and grouped into the same surface, thus bounding the overall number of surfaces.

For a surface $\sigma_{\bf t,\ff}$ and its rounded version $\sigma_{\bf s,\g}$ we have, for each $j$,
\begin{align*}
\left| \left( F_j(\bf x;\bf t)+f_j\right) - \left(F_j(\bf x;\bf s)+g_j\right) \right|
 & \le \eps ,
\end{align*}
where $\eps$ is some parameter depends on the approximation.
The canonization approximates the input set, but significantly reduces the number of surfaces to process and thus runtime.
During canonization we keep track of merged surfaces to recover the original surfaces (the set of inliers) after finding the maximum.

For each surface we first translate the surface such that the minimum box vertex coincides with the parameter space origin.
Recalling the surface definition from Section~\ref{inc_to_surf}, we round each free parameter to the closest integral multiple of $\sfrac{\eps'}{(\ell+1)}$ and each essential parameter to the nearest integer multiple of $\sfrac{\eps'}{(\ell+1)\delta}$ where $\delta$ is the diameter of the current box and $\eps'$ is a constant depending on the given approximation parameter $\eps$ and the surface.

\begin{figure}[t]
    \centering
    \includegraphics[width=8cm]{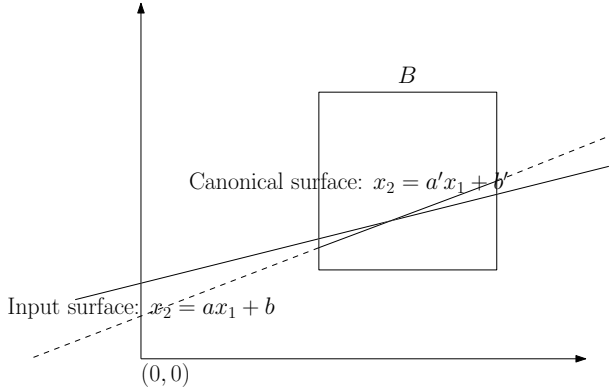}
    \caption{Canonization where the surface is a line in 2D}
    \label{fig:rounding}
\end{figure}

Figure~\ref{fig:rounding} shows the rounding process for the case of a 2D surface given by $x_2=ax_1+b$ and a box $B$.
Note that our goal is to find a surface which is close to the input surface \textit{within the box $B$}.
We first round the essential parameter $a$ to $a'$ which moves the line away from the box, so we translate it by changing $b$ and then rounding it to $b'$.
The key is that the number of canonical surfaces in $B$ is upper bounded independent of the number of input surfaces.

Canonization accumulates error with each rounding (bounded by $\log{1/\eps}$ octree levels).
To guarantee approximation errors we adjust the original $\eps$ to $\eps'=\frac{\eps}{c\log{1/\eps}}$ ($c$ is a some constant depending on the surface, see~\cite{AKKSZ19}).
$\eps'$ is a constant depending on the bounded and Lipschitz gradients of the surfaces, which can be estimated analytically~\cite{WZ96}.

\vspace{\baselineskip}

\noindent\textbf{Estimating $\eps'$ in practice.} $\eps'$ determines the runtime upper bound and the best parameter can vary among coordinates such that in practice we tune constants per parameter (different $\eps$ for $a$ and $b$ in the line fitting example).

\vspace{\baselineskip}

\noindent\textbf{Surface-Box intersection.} There exist a general algorithm for any surface and box: Suppose the surface $\sigma$ is given as a polynomial equation $F(\xx)=0$ and that the box is $B=[0,1]^3$.
For simplicity let us focus on $d=3$, so the surface is $F(x_1,x_2,x_3)=0$.
If we assume $\sigma$ is connected, then $\sigma$ intersects $B$ if either (1) $\sigma$ is fully contained in $B$, or (2) it intersects some face, or (3) it intersects some edge.
To test for case (1), pick an arbitrary point on $\sigma$ and test if it lies in $B$.
If $\sigma$ is not connected, repeat this for one point in each connected component.
Cases (2) and (3) are handled recursively: for (2), we handle each face so that intersect with a 2D plane, where the polynomial is $F(x_1,x_2,0)=0$.
For (3) we have a univariate polynomial $F(x_1,0,0)=0$ and we need to test if it has a root in $[0,1]$.

The above general algorithm can be slow in practice for many dimensions but we found that using a subset of the conditions is a good approximation in practice.

\subsection{Theoretical Analysis of Related Work}

\noindent To the best of our knowledge three alternatives to our algorithm exist and for which we compare asymptotic runtime. 

\vspace{\baselineskip}
\noindent\textbf{RANSAC}~\cite{FB81} is the most common approach to solve outlier rejection problems.
The RANSAC complexity is $O(\frac{n}{\log{(1-b^k)}})$ for a lower bound $b$ on the fraction of inliers, a minimal set size $k$ needed to define a possible solution and $n$ input constraints.
This is linear in $n$ only when $b$ is a constant and even then it grows quickly when $b$ is decreasing, making RANSAC unattractive for large scale problems.

\vspace{\baselineskip}
\noindent\textbf{Hough based methods}~\cite{HOU62,DH72} exist in many variants, but all share a term with polynomial runtime complexity in which all sets of points (each of minimal size $k$) are traversed.
For each subset we vote for the parameters defined by the subset, taking $O(n^k)$ time to find the parameters with the maximum number of votes.
Alternatively for each single point $p$ one can enumerate all bins in the Hough space corresponding to parameters of models passing through (or close) to $p$.
This takes $O(n/\eps^s)$ where $s$ is the dimension of the Hough surface.
A randomized version of Hough voting is applicable, if a lower bound on the number of inliers exists, though, it introduces the same limitation as RANSAC.

\vspace{\baselineskip}
\noindent\textbf{Branch\&Bound}~\cite{HD60} methods have been considered and implemented for a large number of optimization problems, including the family we consider here~\cite{B03,LH07,OKO09,FLOEK16,FL16}.
They are optimal up to an error bound defined by the smallest box that terminates the process (say of size $\eps$ assuming w.l.o.g. that the space is $[0,1]^d$).
The practical runtime of Branch\&Bound can be low but is highly dependent on the structure of the surfaces and the quality of the bounds which are notoriously hard to find.
The worst case runtime is $O(n/\eps^k)$ (ignoring the log factor), where $k$ depends on the dimension of the problem, and is thus similar to the naive enumeration of cells.
Recently, the 6DOF posing problem was solved without correspondences using B\&B~\cite{CPKH20}, against which we evaluate in Section~\ref{sec:gopac-comparison} and in the supplemental material.
The worst case runtime presented in~\cite{CPKH20} is $O(\mu^{-3}\eta^{-6}nm)$ for $m$ 3d points, $n$ frame key points and $\mu,\eta$ are tolerance parameters similar to our $\eps$.
This aligns with the naive method, as $nm$ is the number of matches in the problem (though it can be much faster in practice).

\vspace{\baselineskip}
Our generalized voting approach is asymptotically and in practice faster and more general than these alternatives while being easy to implement.
Existing B\&B algorithms can be combined with the proposed canonization, to leverage high quality bounds in combination with a guaranteed worst case runtime.

\section{Applications}

In the following, we enumerate a list of typical outlier rejection problems in geometric computer vision and we show how each one of them can be reduced to the incidence problem and solved efficiently using our method.
We deliberately do not compare to the numerous state of the art methods across all applications, but rather want to emphasize the \textit{generality of our approach}.

\subsection{Fitting Hyperplanes in $d$-space}\label{sec:line_fitting}
Model fitting is a well investigated problem in the literature~\cite{SS01,DH72,HOU62}, and commonly solved using RANSAC or Hough voting. It's also mentioned as an example in~\cite{AKKSZ19} and solved using a primal-dual method.
The goal is to report the hyperplane consistent with most points (distance $\leq \eps$).

In order to solve this using our general method, we first use the point-hyperplane duality~\cite{EA57}.
Points are transformed to hyperplanes and hyperplanes to points with the vertical distance between these plane-point pair being preserved.
If we keep hyperplanes in appropriate orientation this algebraic distance (the $d$ coordinate) is a good approximation to the Euclidean distance we want to minimize.
We then search for the point that is consistent with the maximum number of hyperplanes (=surfaces) and transform it back via duality to obtain the best fitting hyperplane.
For example the parameterization in 3D is the standard plane equation:
\begin{equation} \label{planes}
x_d=a_0+\Sigma^{d-1}_{i=1} x_i a_i.
\end{equation}
with $\ell=d-1,k=d-1$ in $R^d$ and the runtime is $O(n+\frac{\polylog (1/\eps)}{\eps^d})$ compared to $O(n/\eps^{d-1})$ in the naive method.
The $d$-dimensional algorithm is then better for any $n>1/\eps$ (for any fixed $d$).

\begin{figure}[t]
    \centering
    \includegraphics[width=8.0cm]{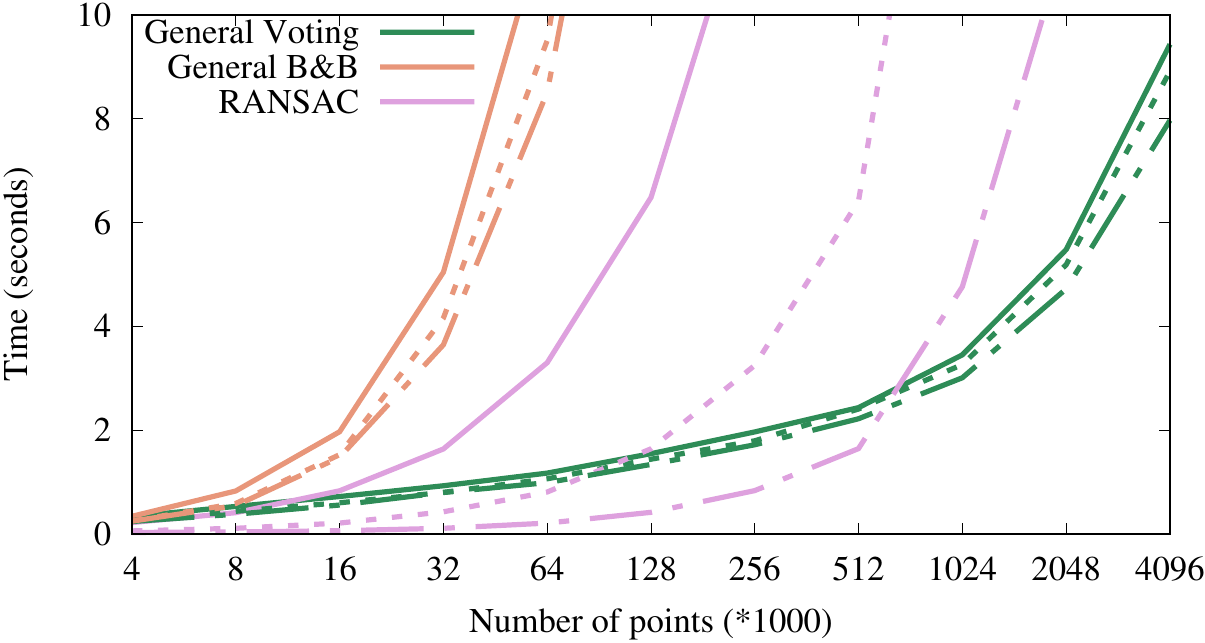}
    \caption{Runtime comparison for line fitting with various inlier fractions: solid 1\%, dotted 2\%, dashed 4\%.}
    \label{fig:line_fitting}
\end{figure}

\paragraph{Evaluation}
We evaluate the runtime for the 2D case by comparing B\&B and Ransac to our method where we implemented the problem specific surface definition and the intersection predicate.
The test data consists of uniformly sampled outlier points in the unit cube and inlier points sampled along a fixed line with additive noise.
We apply this for both, an increasing number of points and increasing fraction of outliers (which we provide to Ransac to estimate the iteration count).
As can be seen in Fig.~\ref{fig:line_fitting} for small inlier fractions Ransac becomes slow and our method outperforms due to the better asymptotic runtime.
Results for B\&B represent the worst case due to the uniformity of the input data.
In all cases, accurate line parameter were found using $\eps=0.002$.

\subsection{Absolute Camera Posing Problems}
\label{sec:camera_posing}
The authors of~\cite{AKKSZ19} show how to define the surfaces for a calibrated camera with known gravity direction which is a setup with 4 DoF that is also used in~\cite{ZSP15,SEKO16}.
In the following we derive formulations for camera setups with higher DoF that are difficult to handle by a naive voting approach due to the higher dimensional solution space.
Our generalized voting provides a good approximation of the best pose which can then be taken by a more accurate method (e.g.\ a minimal solver within RANSAC) that then operates on the remaining point set largely free from outliers.

Let $\ww =(w_1,w_2,w_3)$ be a point in $\reals^3$ and let $(\xi,\eta)$ be a point in the (normalized) image plane. The triple $(\ww,\xi,\eta)$ is a correspondence $\cc$ and we show how the constraints from the correspondence are reduced to a general surface $\sigma_\cc$.
We provide runtime complexities of the following formulations in Table~\ref{tab:camera-posing-complexities}.

\begin{center}
\begin{table*}[t]
\small
    \centering
    \begin{tabular}{l l l c c l}
        \toprule
        Posing problem & Known focal & Known gravity & Generalized voting & Naive voting & General faster naive for \\
        \midrule
        4DoF~\cite{AKKSZ19}  & \checkmark & \checkmark & $\tilde{O}\left(n+1 / \eps^6\right)$ & $O(n/\eps^2)$ & $n > 1/\eps^4$\\
        5DoF                 & - & \checkmark & $\tilde{O}\left(n+1 / \eps^6\right)$ & $O(n/\eps^3)$ & $n>1/\eps^3$ \\
        6DoF                 & \checkmark & - &  $\tilde{O}\left(n+1 / \eps^7\right)$ & $O(n/\eps^4)$         &   $n>1/\eps^3$       \\
        7DoF                 & - & - & $\tilde{O}\left(n+1 / \eps^7\right)$ & $O(n/\eps^5)$ & $n>1/\eps^2$ \\
        5DoF (radial camera) & - & - & $\tilde{O}\left(n+1 / \eps^6\right)$ & $O(n/\eps^4)$ & $n>1/\eps^2$ \\
        \bottomrule
    \end{tabular}
    \caption{Complexity analysis for the camera posing formulations of Sec.~\ref{sec:camera_posing}. We use the definition $\tilde{O}(n+1/f(\eps)) = O(n+\polylog(1/\eps)/f(\eps))$ for simplicity.}
    \label{tab:camera-posing-complexities}
\end{table*}
\end{center}

\subsubsection{Unknown focal length, known gravity direction (5DoF)}
\label{5DoF_focal}
We use $\reals^5$ with coordinates $(x,y,z,\kappa,f)$ as our search space, where $(x,y,z)$ is the camera position, $\kappa=\tan\theta$ with $\theta$ denoting the (remaining) camera orientation around gravity, and $f$ is the focal length.
Each such coordinate models a possible pose of the camera plus focal length.
A correspondence $\cc$ is parameterized by the triple $(\ww,\xi,\eta)$, and defines a $3$-dimensional algebraic surface $\sigma_\cc$.
It is the locus of all camera poses and focal lengths at which it sees $w$ at image coordinates $(\xi f,\eta f)$.
We can rewrite these equations into the following parametric representation of $\sigma_\cc$,
expressing $z$ and $\kappa$ as functions of $x$,$y$ and $f$.
%
%
\begin{align} \label{cparw}
& \kappa  = \frac{(w_2-y) - \xi f (w_1-x)} {(w_1-x) + \xi f (w_2-y)}  \\
& z  = w_3 - \eta f \sqrt{(w_1-x)^2+(w_2-y)^2}
\end{align}

Our goal is to find a tuple $(x,y,z,\kappa,f)$ such that as many correspondences as possible are (approximately) \emph{consistent} with it.
In other words the ray from the camera
center $c$ to $w$ goes approximately (i.e.\ up to some reprojection error) through $(\xi f, \eta f)$ in the image plane.

In the case of our $3$-surfaces in $5$-space, the parameter $w_3$ is free, and we introduce a second artificial free parameter into equation~\ref{cparw} for $\kappa$.%
The number of essential parameters is $\ell=4$ (they are $w_1$,$w_2$,$\xi$, and $\eta$).
With $d=5$ and $k=3$, using the general technique from~\cite{AKKSZ19}, we obtain an algorithm that takes (for $n$ correspondences) $O(n+\frac{\polylog (1/\eps)}{\eps^6})$.
The naive method would take $O(n/\eps^3)$ time so that the general technique is asymptotically better (ignoring poly logarithmic factors) for any $n>1/\eps^3$.

\subsubsection{Unknown focal length (7DoF)}
\label{7DOF}

We use $(x,y,z,\boldsymbol{\phi},f)$ as the $7$-tuple of unknowns that we aim to solve for, where $\boldsymbol{\phi}$ is a 3-vector describing the minimal rotation parameterization (e.g. angle-axis).
Given the standard general $3 \times 4$ projection matrix 
$P =  \mathrm{diag}(f,f,1) \begin{bmatrix} R(\boldsymbol{\phi}) & \bf{t} \end{bmatrix} $
%
the world point $\ww$ projects to 
\begin{equation}
\lambda \begin{pmatrix} \xi \\ \eta \\ 1 \end{pmatrix} = P \begin{pmatrix}\ww \\ 1 \end{pmatrix} = \begin{pmatrix} f ({\bf r_1} \ww + x) \\ f ({\bf r_2} \ww + y) \\ {\bf r_3} \ww + z \end{pmatrix} ,
\end{equation}
where $\bf{r_i}$ denote the $i^{th}$ row of the rotation matrix.
This reveals constraints 
\begin{equation}
\begin{aligned}
0 &= f ({\bf r_1} \ww + x) - \xi ({\bf r_3} \ww + z) \\
0 &= f ({\bf r_2} \ww + y) -\eta ({\bf r_3} \ww + z)
\end{aligned}
\end{equation}
which allows to parameterize $x$ and $y$ as a a function of $f,\boldsymbol{\phi},z$ for each correspondence $(\ww,\xi,\eta)$ according to
\begin{equation}
\begin{aligned}
x&=\xi({\bf r_3} \ww + z) / f - {\bf r_1} \ww \\
y&=\eta({\bf r_3} \ww + z) / f - {\bf r_2} \ww
\end{aligned}
\end{equation}

With $d=7,\ell=5,k=5$ the naive rendering takes $O(n/\eps^5)$ while general voting takes $O(n+\frac{\polylog (1/\eps)}{\eps^7})$ which is better for any $n>1/\eps^2$.

\subsubsection{Calibrated camera (6DoF)}
\label{6DOF}
This is identical to the 7DoF case where we set $f=1$.
With $d=6,\ell=5,k=4$, naive rendering takes $O(n/\eps^4)$ while generalized voting takes $O(n+\frac{\polylog (1/\eps)}{\eps^7})$ as for 7DoF case which is better for any $n>1/\eps^3$.

\subsubsection{Unknown focal length using a radial camera model (5DoF)}
\label{radial}
The previous formulation for the 6DoF pose plus focal length case requires a $7$-dimensional voting space.
As an alternative we propose to leverage a radial camera model~\cite{TP12} which is known to work well for pose estimation with unknown focal length~\cite{BZT10,LSKP19}.
The approach factors the pose estimation in two consecutive steps where the first relies on line-to-point correspondences and solves for all parameters except the focal length and the camera motion along the optical axis.
A second \emph{upgrade} step then solves for the remaining parameters using a least squares fit.
For outlier removal we focus on the first step and therefore are looking for a $5$-tuple $(x,y,\boldsymbol{\phi})$.
The $2 \times 4$ radial camera projection matrix is
\begin{equation}
P=\begin{bmatrix}
 {\bf r_1} & -y \\
 {\bf r_2} & x 
\end{bmatrix}
\end{equation}
and we have $[\ww^T,1] (\eta {\bf p_1} - \xi {\bf p_2}) =0$, where ${\bf p_1},{\bf p_2}$ are the rows of $P$.
We then obtain $\sigma_\ww$ where we parameterize $x$ as a function of $y,\boldsymbol{\phi}$:
\begin{equation}
\begin{aligned}
x=\eta({\bf r_1} \ww - y) / \xi - {\bf r_2} \ww.
\end{aligned}
\end{equation}

We have $d=5,\ell=5,k=4$ and the algorithm takes $O(n+\frac{\polylog (1/\eps)}{\eps^6})$ time, compared to $O(n/\eps^4)$ with naive rendering. This makes the general algorithm better for any $n>1/\eps^2$.

\subsubsection{Evaluation}
Localizing an uncalibrated camera with a known axis of rotation is a common problem in computer vision and both, the problem dimensionality and the number of unknowns are well suited to demonstrate the general use of our method. 
Therefore, we focus on the 5DOF problem described in Section~\ref{5DoF_focal} here and compare the runtimes of our generalized voting approach with naive voting, Branch\&Bound, and RANSAC.
Note, that Section~\ref{sec:gopac-comparison} presents a more in depth evaluation, including localization performance, against the start-of-the-art (Branch\&Bound) method for 6DoF camera posing on public datasets.

Our evaluation data exhibits real-world, large-scale scenes where a set of $3$d points and potentially corresponding image points are given.
We create them by matching a real query image against 3D points of a SfM model. Using 7, 14, and 56 candidates in the nearest neighbor search results in problem sizes of $\sim10k$, $\sim15k$, and $\sim20k$ matches.

For the solution computations we consider typical space limitations for all methods:
Due to the nature of the parameterization we only consider a camera orientation with tangent in $[-1,1]$, rotating the scene accordingly if needed. We bound the spatial position to be at most $50$m and the camera height $5$m from the ground truth.
The focal length is bounded within a fraction $[0.6, 1.3]$ of the ground truth.

\begin{figure}[t]
    \centering
    \includegraphics[width=4.1cm]{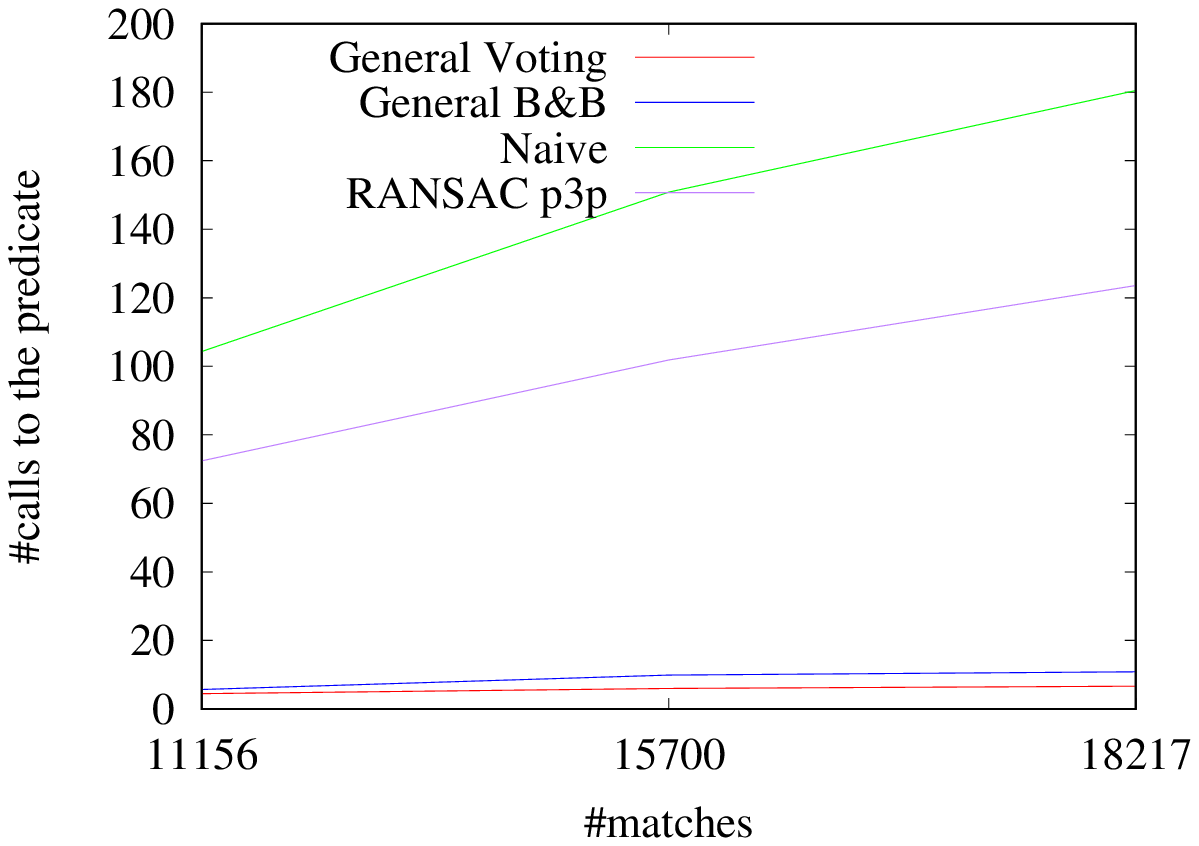}
    \includegraphics[width=4.1cm]{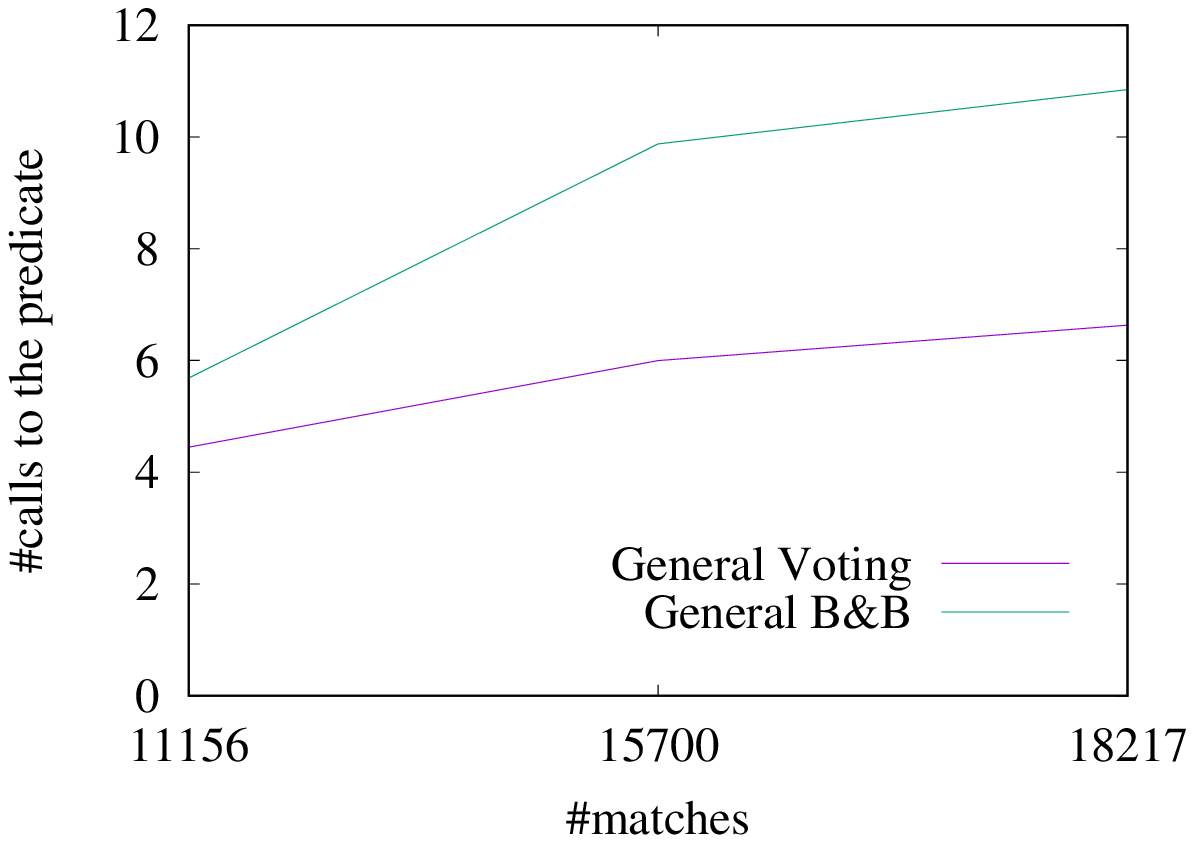}
    \caption{Runtime comparison of our generalized voting approach against alternatives in dependence on the number of matches for camera posing: (left) demonstrates the high cost of naive voting, while (right) provides a better comparison between B\&B and our generalized voting.}
    \label{fig:real_data_time}
\end{figure}

Figure~\ref{fig:real_data_time} illustrates our results which demonstrate that our method is the most efficient. 
In order to eliminate implementation details which can change runtimes  considerably, we do not measure time but instead count the number of dominant operations.
That is the number of $5$d grid cells that are touched in the naive implementation and the number of calls to the surface-box intersection predicate for our algorithm and for B\&B. 
For RANSAC we simplified the problem to one with known focal length and were thus able to use a 3-point minimal solver with early rejection to account for the known gravity direction. We also tuned the number of iterations and inlier tolerance to find a good pose with minimal runtime.
As expected, the gap between the methods and the improvement in the proposed algorithm is increasing with the size of the input as suggested by the asymmetrical complexity.

\subsection{Intersections of Rays/Lines in $3$-space}
The problem setup consists of a set of input $n$ rays in $3$-space and a grid of cell size $\eps$ (within some bounding box).
The goal is to report all subsets of rays intersecting any (non-empty) cell and with cardinality larger than some threshold.
We can formulate the general surfaces using:
\begin{equation} \label{lines}
y  = ax+b  \quad\quad\quad
z  = cx+d.
\end{equation}
With $\ell=2,k=1,d=3$ the naive algorithm takes $O(n/\eps)$ while the general voting takes $O(n+\frac{\polylog (1/\eps)}{\eps^4})$ which is better (ignoring logarithmic factors) for any $n>1/\eps^3$.

\subsection{Pointset Alignment}
In this problem $A$ and $B$ are sets of points in the plane.
Any hypothetical correspondence between $p \in A$ and $q \in B$ defines a surface in the $4$-space of similarity transformations (translation, rotation and scale).
The linear parametrization is:
\begin{align}
&q_x=ap_x+bp_y+c \\
&q_y=-bp_x+ap_y+d
\end{align}
where $a=s\cos(\theta),b=s\sin(\theta),s$ is the scale, $\theta$ is the rotation and $(c,d)$ is the translation vector. 
The general surface is then:
\begin{align}
& a= -\frac{c p_x+d p_y-p_x q_x-p_y q_y}{p_x^2+p_y^2} \\
& b= -\frac{c p_y-d p_x-p_y q_x+p_x q_y}{p_x^2+p_y^2}.
\end{align}

Which is a $2$-surface embedded in $R^4$ where $a,b$ are given as a functions of $c,d$.
In this case we have two essential parameters ($c,d$) and we introduce two artificial new free parameters.
With $\ell=2,k=2$ and $d=4$ with $n$ as the number of canonical surfaces the naive algorithm would take $O(\frac{n}{\eps^2})$ and the general voting takes $O(n+\frac{1}{\eps^4}\log{\frac{1}{\eps}})$ which is better for any $n>\frac{1}{\eps^2}$ (ignoring logarithmic factors).

\section{Comparison with GOPAC~\cite{CPKH20}}
\label{sec:gopac-comparison}
Solving the 6DOF camera posing problem without known correspondences is hard due to the large search space.
Recently, the authors of~\cite{CPKH20} presented a solution using a globally optimal method based on Branch\&Bound and conducted extensive evaluations on public datasets (Data61/2D3D~\cite{NNSP15} and Stanford 2D-3D-Semantics (2D-3D-S)~\cite{ASZS17}) against state-of-the-art alternatives.

We evaluate our general voting on this problem, viewing it as a Branch\&Bound search. We compare our results with~\cite{CPKH20} on the same datasets, using the GOPAC code provided by the authors (re-run on our machine to ensure comparability to our approach), showing that our general algorithm is not only asymptotically better but also faster in practice. Both, our algorithm and GOPAC, are globally optimal up to a prescribed resolution $\eps$ and perform joint inlier set maximization and correspondence search. 
Our method achieves a significantly better worst case runtime of $O(n+\poly(1/\eps))$ compared to $O(n*\poly(1/\eps))$ in GOPAC.
%

In order to implement the 6DOF solver in our general framework we reformulated the problem as a surface consensus problem according to Sec.~\ref{6DOF}.
Compared to GOPAC our formulation does not use an angular error metric, but is based on surface distances. However, it is possible to obtain the globally optimal solution by conservative expansion of the cubes (to avoid missing inliers) and verify the angular projection error on the final inlier set for the minimal cuboids.

\paragraph{Data61/2D3D~\cite{NNSP15} outdoor dataset:}
The dataset consists of a 88 3D points and 11 sets of 30 bearing vectors.
Table~\ref{tab:posing-performance-outdoor} shows the localization performance and runtime for GOPAC, our generalized voting and RANSAC.
%
%
Both Branch\&Bound algorithms show comparable accuracy due to their global optimality and we only expect a difference in runtime. Here our algorithm is more than an order of magnitude faster.
Equivalent to~\cite{CPKH20} we restrict the translation solution to a $50m \times 5m \times 5m$ domain along the street (as it's known that cameras are mounted on a vehicle), and a camera is considered successfully posed, if the rotation error is less than $0.1$ radians and the normalized translation error is less than $0.1$.

\begin{table}[ht]
\begin{center}
 \begin{tabular}{llll}
 \toprule 
 Algorithm & GOPAC & GV (ours) & RANSAC\\
 \midrule
 Translation Error (m) & 2.76 & 2.89 & 28.5\\
 Rotation Error ($\deg$) & 2.18 & 0.46 &179\\
 Runtime (s) & 457 & 27 & 422\\
 Success rate (inliers) & 1 & 1 & 0\\
 Success rate (pose)& 0.82 & 0.82 & 0.09 \\
 \bottomrule
\end{tabular}
\end{center}
\caption{Camera posing results (median error over the 11 queries) for Scene 1 of the Data61/2D3D dataset. Error metrics for GOPAC are taken from~\cite{CPKH20}). Runtimes are from single-threaded execution of C++ code, where GOPAC was rerun on our machine.}
\label{tab:posing-performance-outdoor}
\end{table}

\begin{figure}[t]
    \centering
    \includegraphics[width=8cm]{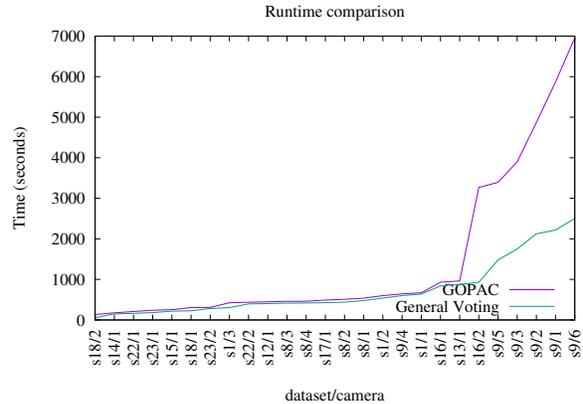}
    \caption{Sorted runtimes for all queries in the Stanford 2D-3D-S dataset. Same experiment as Table~\ref{tab:posing-peroformance-indoor}.}
    \label{fig:query-runtimes}
\end{figure}

\paragraph{Stanford 2D-3D-Semantics~\cite{ASZS17} indoor dataset:}
The dataset consists of 15 rooms of different types and 27 sets of 50 bearing vectors.
Table~\ref{tab:posing-peroformance-indoor} lists our performance and runtime comparison.
As the experimental evaluation of~\cite{CPKH20} uses a GPU implementation we reran both algorithms on CPU using 8 threads to ensure comparability. In order to obtain reasonable runtimes we set $\eps$ to 0.5m and 0.1 radians, respectively for both methods.
Again we expect a similar localization performance due to the global optimality of both methods. On average the runtime of our algorithm on this dataset is only slightly better.
As noted in Section~\ref{algs}, the effect of the asymptotically better worst case runtime increases with the hardness of the problem (both, in size and data distribution). 
Therefore, Figure~\ref{fig:query-runtimes} depicts the (sorted) runtimes for all queries, which illustrates that our algorithm becomes significantly faster for the hard cases.

\begin{table}[ht]
\begin{center}
 \begin{tabular}{lll}
 \toprule 
 Algorithm & GOPAC & GV (ours) \\
 \midrule
 Translation Error (m) & 0.29 & 0.38 \\
 Rotation Error ($\deg$) & 3.46 & 2.81 \\
 Runtime (s) & 508 & 421 \\
 Success rate (inliers) & 1 & 1 \\
 Success rate (pose) & 0.77 & 0.77\\
 Success rate within 60s & 0.19 & 0.42

\\bottomrule
\end{tabular}
\end{center}
\caption{Camera posing results (median error) for Area 3 of the Stanford 2D-3D-S dataset (see Tab. 3 in~\cite{CPKH20}). Runtimes are from running C++ code on CPU with 8-threads.}
\label{tab:posing-peroformance-indoor}
\end{table}

The supplementary material contains additional comparisons on large scale outdoor datasets.

\section{Conclusion}


In this paper we introduced the concept of \textit{general voting}, a powerful method for outlier rejection applicable to a wide range of geometric computer vision problems.
We adopt the previously proposed method of \textit{approximate incidences} to solve for inlier maximization in multiple classical computer vision problems ranging from camera posing and ray intersection to geometric model fitting.
We described the general recipe with a simple to understand 2d-line fitting example, but also demonstrated its applicability to real-world problems.

Through theoretical analysis and experiments we demonstrated that our algorithm scales better than its alternatives, like RANSAC or Branch\&Bound, both in terms of complexity and real-world runtime.
The experimental data validated that the proposed solution performs particularly well for large problems with low inlier ratios where alternative solutions require problem specific knowledge to remain applicable.

One of the key use-cases we investigated is camera posing with and without known gravity direction and focal length, problems that cannot be solved efficiently at large scale with previously published methods, yet that have wide applicability in industry.
To solve these cases, we introduced two algorithms that are key to efficiency: canonization of the intersected surfaces and an efficient $d$-box intersection algorithm which we combine in a spatial subdivision scheme.
To demonstrate the impact of these contributions we provided an extensive evaluation against a recently published state-of-the art method on publicly available, large-scale indoor and outdoor datasets.

Beside solving concrete localization approaches this work introduced the concept of general voting to the wider computer-vision community.
We aim to provide a recipe for applying this approach to a range of problems and publish an open-source implementation of our efficient general voting to unlock new applications and research directions. 

\paragraph{Acknowledgements}
The authors would like to thank Micha Sharir for helpful discussions concerning the general surface-box intersection.

\clearpage 

{\small
\bibliographystyle{ieee_fullname}
\bibliography{egbib}
}

\end{document}